\documentclass[10pt,twocolumn,letterpaper]{article}

\usepackage{wacv}              

\usepackage[accsupp]{axessibility}
\usepackage{amsmath}
\usepackage{amssymb}
\usepackage{booktabs}

\usepackage{subcaption}
\usepackage{makecell}

\usepackage{mathtools}
\usepackage{amsthm}

\theoremstyle{plain}
\newtheorem{mytheorem}{Theorem}[section]

\theoremstyle{remark}
\newtheorem{remark}[mytheorem]{Remark}

\usepackage[textsize=tiny]{todonotes}
\setuptodonotes{inline}

\usepackage[utf8]{inputenc}
\usepackage{tikz-cd}
\tikzcdset{scale cd/.style={every label/.append style={scale=#1}, cells={nodes={scale=#1}}}}
\usetikzlibrary{mindmap, backgrounds, calc}

\usepackage{nicefrac}
\usepackage{multirow}

\newcommand{\R}{\ensuremath{\mathbb{R}}}

\DeclareMathOperator{\sign}{sign}
\DeclareMathOperator{\Var}{Var}

\DeclareMathOperator*{\argmin}{arg\,min}

\usepackage[pagebackref,breaklinks,colorlinks]{hyperref}

\usepackage[capitalize]{cleveref}
\crefname{section}{Sec.}{Secs.}
\Crefname{section}{Section}{Sections}
\Crefname{table}{Table}{Tables}
\crefname{table}{Tab.}{Tabs.}

\def\wacvPaperID{2332} 
\def\confName{WACV}
\def\confYear{2024}

\begin{document}

\title{Increasing biases can be more efficient than increasing weights}

\author{
Carlo Metta$^0$\\
ISTI-CNR Pisa, Italy\\
\and
Marco Fantozzi\\
University of Parma, Italy\\
\and
Andrea Papini\\
Scuola Normale Superiore, Pisa, Italy\\
\and
Gianluca Amato$^1$\\
University of Chieti-Pescara, Italy\\
\and
Matteo Bergamaschi\\
University of Padova, Italy\\
\and
Silvia Giulia Galfrè$^2$\\
University of Pisa, Italy\\
\and
Alessandro Marchetti$^3$\\
University of Chieti-Pescara, Italy\\
\and
Michelangelo Vegliò\\
University of Chieti-Pescara, Italy\\
\and
Maurizio Parton$^1$\\
University of Chieti-Pescara, Italy\\
\and
Francesco Morandin$^1$\\
University of Parma, Italy\\
}
\maketitle


\begin{abstract}

We introduce a novel computational unit for neural networks that features multiple biases, challenging the traditional perceptron structure. This unit emphasizes the importance of preserving uncorrupted information as it is passed from one unit to the next, applying activation functions later in the process with specialized biases for each unit. Through both empirical and theoretical analyses, we show that by focusing on increasing biases rather than weights, there is potential for significant enhancement in a neural network model's performance. This approach offers an alternative perspective on optimizing information flow within neural networks. See source code~\cite{CurDAC}.
\end{abstract}

\section{Introduction}
\label{sec:introduction}

\footnotetext[0]{EU Horizon 2020: G.A. 871042 SoBig-Data++, NextGenEU - PNRR-PEAI (M4C2, investment 1.3) FAIR and “SoBigData.it”.}
\footnotetext[1]{Funded by INdAM (groups GNAMPA, GNCS, and GNSAGA).}
\footnotetext[2]{Partially supported by SPARK Pisa.}
\footnotetext[3]{National PhD in AI, XXXVII cycle, health and life sciences, UCBM.}

\let\oldthefootnote=\thefootnote
\renewcommand{\thefootnote}{}
\footnotetext{$^\text{all}$Computational resources provided by CLAI lab, Chieti-Pescara.}
\let\thefootnote=\oldthefootnote

Historically the structure of the perceptron, the artificial neural
network's fundamental computational unit, has rarely been questioned.
The biological inspiration is straightforward: input signals from the
dendrites are accumulated at the soma (with a linear combination), and
if the result is above the activation threshold (that is, the opposite
of some bias) there is a nonlinear reaction, as the neuron fires along
the axon (with the activation function).

In time, the early sigmoid activation function was replaced by ReLU
and variants, and the biological analogy became less stringent,
shifting focus on the desirable mathematical properties of the class
of functions computed by the networks, like representation power and
non-vanishing gradients.

This has brought us to the current situation in which most units
output their signal through a nonlinear activation function which
effectively destroys some information.  In fact, ReLU is not
invertible, as it collapses to zero all negative values.  Though some
of its variants may be formally invertible (leaky ReLU~\cite{MHNLRL}
and ELU~\cite{CUHELU} for example), the fact that they overall perform
in a way very similar to ReLU, suggests that their way of compressing
negative values via a small derivative bijection leads to the same
general properties of the latter.

In this paper, we investigate a radical rethinking of the standard
computational unit, where the output brings its full, uncorrupted
information to the next units, and only at this point is the
activation function applied, with biases specialized for each unit.
From the biological point of view, this is like having the activation
at the dendrites instead of at the base of the axon, and
correspondingly we call the new unit `DAC', for `Dendrite-Activated
Connection'.

This kind of reversed view has already proved fruitful in the
evolution from ResNets v1~\cite{HZRDRL} to v2~\cite{HZRIMD} when a
comprehensive ablation study showed that for residual networks it is
better to keep the information backbone free of activations for
maximum information propagation, and pre-activate the convolutional
layers in the residual branch.

Here this view is taken forward: not only the units are pre-activated,
but the biases become specific to each input-output pair, as the
weights are.  We refer to this as having \emph{unshared biases}.



The main result of this paper is evidence that sometimes incorporating more biases can increase accuracy more than adding weights, without altering model complexity (see Section~\ref{sec:experiments}, SGEMM subsection). The fact that DAC consistently outperforms the baseline models across diverse architectures and datasets strengthens this finding, see the rest of Section~\ref{sec:experiments}. Hence, DAC emerges as a viable strategy for enhancing neural network performance when increasing weights proves ineffective.

The proposed model is introduced in Section~\ref{sec:dac_idea} with a
first theoretical discussion.  Related works are listed in
Section~\ref{sec:related}.  The main practical details for replacing
shared with unshared biases are discussed in
Section~\ref{sec:methods}, and the empirical experiments can be found
in Section~\ref{sec:experiments}.  Finally, further theoretical
questions can be found in Section~\ref{sec:theoretical_main} and in
the Appendix.


\section{Model}
\label{sec:dac_idea}

In a standard neural network, units are often post-activated: they
compute a linear combination of their inputs and then apply a
nonlinear filter to the result.
\begin{equation}
\label{eq:standard}
\begin{cases}
    z_i = \sum_{j\in\mathcal I_i} w_{i,j}\,y_j & \quad \text{linear aggregation}\\
    y_i = \varphi(b_i + z_i)  & \quad \text{nonlinear filter}
\end{cases}    
\end{equation}
where, for unit $i$, $\mathcal{I}_i$ denotes the set of input nodes,
$b_i$ the bias and $\varphi$ the activation function.

We propose to consider units that are pre-activated with unshared
biases:
\begin{equation}
\label{eq:dac_equation}
\begin{cases}
    y_{i,j} = \varphi(b_{i,j} + z_j)  & \quad \text{nonlinear filter}\\
    z_i = \sum_{j\in\mathcal I_i} w_{i,j}\,y_{i,j} & \quad \text{linear aggregation}
\end{cases}    
\end{equation}

\begin{figure}[t]
    \centering
    \includegraphics[scale=0.5]{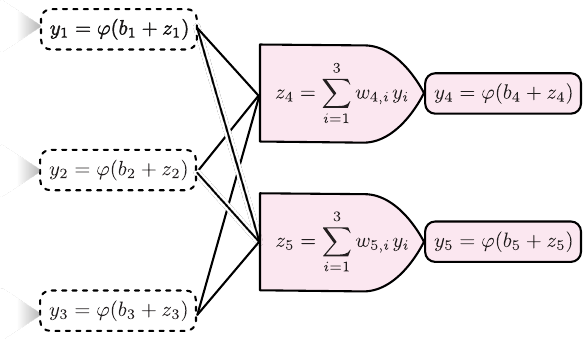}
    \caption{Standard connection between two consecutive layers.  The
      output layer (pink) is fully connected and has two units labelled 4 and
      5.  The input layer has three units:
      $\mathcal{I}_4=\mathcal{I}_5=\{1,2,3\}$.  Bullets and rectangles
      represent linear aggregation and nonlinear filters
      from~\eqref{eq:standard}, respectively.  Units 4 and 5 must share
      the same biases $b_1,b_2,b_3$ in the activation of their inputs.}
    \label{fig:classical_unit}
\end{figure}

\begin{figure}[t]
    \centering
    \includegraphics[scale=0.5]{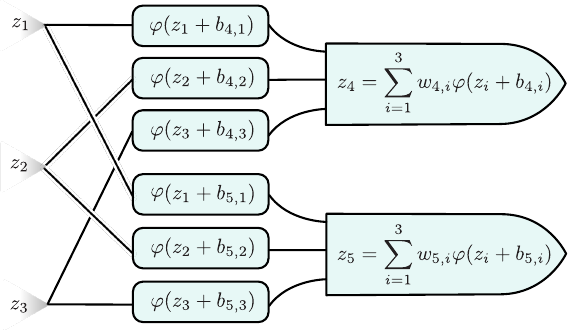}
    \caption{Same structure as in Figure~\ref{fig:classical_unit} with
      post-activation replaced by pre-activation with unshared biases.
      Rectangles and bullets represent nonlinear filters and linear
      aggregations from~\eqref{eq:dac_equation}, respectively.  The
      biases in the activation between the input and the output layer
      depend both on the input node (1, 2 or 3) and the output node (4
      or 5), and so, from the point of view of the output units, we
      refer to them as \emph{unshared}.}
    \label{fig:dac_unit}
\end{figure}

In~\eqref{eq:standard} there is one weight $w_{i,j}$ for each
connection between units and one bias $b_i$ for each unit.
In~\eqref{eq:dac_equation} each connection still has its own weight
$w_{i,j}$, but it also has its own nonlinear filter with a specific
bias $b_{i,j}$.  (Compare Figures~\ref{fig:classical_unit}
and~\ref{fig:dac_unit}.)

Connections between units correspond to synapses or dendrites in
biological neurons, and it is known that activation at the level of
dendrites can actually occur in biological neural
networks~\cite{LarADC,MagDIE}.  With this motivation we refer to a
connection as in~\eqref{eq:dac_equation} as a Dendrite-Activated
Connection (DAC).  See Section~\ref{sec:bio_inspiration} in the
Appendix for additional details on the biological inspiration.

A DAC unit is more general than a standard unit, in fact it can have
almost twice as many parameters, and a correspondingly greater
representation capacity (see Section~\ref{sec:representation_power}
of the Appendix on representation power).  For this reason, in
experiments, neural networks with DACs should be compared to standard
ones with similar number of parameters or computational complexity,
and not with the same number of units or channels.  In this regard it
is important to note that, in convolutions larger than $1\times1$, DAC
biases can be partially shared, yielding a much lesser increase in the
number of parameters (see Section~\ref{sec:methods}).

In this paper we investigate the properties of DAC units and the
behaviour of typical network structures when standard connections are
replaced with DACs.

\begin{remark}\label{rm:pre_and_post_activation}
DAC and the standard post-activation may coexist in the same
connection (in analogy with what happens in the biological neuron,
where the standard post-activation of the axon is always present):
\begin{multline*}
\text{linear combination}
\to
\text{shared bias post-activation}\\
\to
\text{unshared biases pre-activation}
\to
\text{linear combination}
\end{multline*}
For ReLU activations, the composition of post-activation and
pre-activation is equivalent to a pre-activation with modified
coefficients, so we do not investigate further this generalization.
\end{remark}

\section{Related work}
\label{sec:related}

The multi-bias activation (MBA) from~\cite{LOWMBA} replicates $K$
times the input features $z_j$, applies a different bias parameter
$b_j^{(k)}$ to each of them, filters them with ReLU, and then computes
a linear combination over $j$ and $k$ for every output node $i$:
\begin{equation}
\label{eq:mba}
z_i
=\sum_{j\in\mathcal I_i}\sum_{k=1}^Kw_{i,j,k}\,\varphi\bigl(b_j^{(k)}+z_j\bigr).    
\end{equation}
Equation~\eqref{eq:mba} resembles our equation~\eqref{eq:dac_equation}
of pre-activated units.  However, in~\eqref{eq:mba} the pre-activation
biases do not depend explicitly on the output $i$, so they are
multiplied in number, but still effectively shared from the point of
view of the linear operator.  Another consequence of this is that
parameters and computations are increased $K$-fold.  Squeeze
MBA~\cite{FLBSMB} is a variation of MBA that still shares biases among
outputs but tweaks the network structure to partially reduce the
number of parameters.

Other approaches to mitigate the loss of information intrinsic in the
ReLU activation, such as Maxout networks~\cite{GWBMON}, adaptive
piecewise linear activation functions~\cite{AHSAPL}, Concatenated
ReLU~\cite{SSLCRL}, and Activation ensembles~\cite{KlHAEN}, generalize
the activation function by using multiple biases (among other
parameters) but they all maintain a single nonlinear filtered output
per node, and hence shared biases.

There are also ways to design a network that \emph{inherently}
mitigate or avoid the loss of information of activations with shared
biases.  ResNet v2 architecture~\cite{HZRIMD} keeps the information
backbone free of nonlinearities that are only on residual branches
(with pre-activation).  All the nonlinear blocks take their inputs
from the backbone, so each input is the pure linear sum of all the
previous nonlinear branches.  ConvNeXt~\cite{LMWACF} not only uses the
same linear backbone as above, but then applies depthwise
convolutions, brought to popularity by~\cite{ChoXDL} and common to other recent successful architectures.  In depthwise
convolutions, every input channel is unique to one output kernel: this
is a simple solution to avoid sharing biases, though the consequent
low capacity requires that depthwise convolutions are used together
with other types of layers that will typically have shared biases.


\section{Methods}
\label{sec:methods}

To use DAC in a given neural network structure, one replaces the usual
post-activations of computational units with pre-activations for the
downstream units, using unshared biases.  To this end one can design
dense and convolutional units that include the required
pre-activation.

For a dense layer with $n$ units and $m$
inputs,~\eqref{eq:dac_equation} becomes:
\begin{equation}\label{eq:dac_dense}
f_i(z)=\sum_{j=1}^mw_{i,j}\,\varphi(b_{i,j}+z_j)
,\qquad
i=1,2,\dots,n
\end{equation}
In the case of a 2d convolutional layer with $n$ units/kernels and $m$
input channels, we get instead:
\begin{multline}\label{eq:dac_conv}
f_{h,k,i}(z)=\sum_{a,b=-l}^l\sum_{j=1}^mw_{a,b,i,j}\,\varphi(b_{i,j}+z_{h+a,k+b,j})
,\\
i=1,2,\dots,n,\quad (h,k)\in\text{grid}
\end{multline}
where $L=2l+1$ and $L\times L$ is the kernels size.

In the latter equation, if one were to follow the principle
in~\eqref{eq:dac_equation}, for which there should be one
pre-activation bias for every weight, then these totally unshared
biases would take the more general form $b_{a,b,i,j}$ depending on
channel, kernel and position in the kernel.  There is however the
possibility, in this case, to partially share biases and have
$b_{i,j}$ depend on input channel and output kernel only, to get a
better trade-off between flexibility and number of parameters.

In most cases, pre-activated layers~\eqref{eq:dac_dense}
and~\eqref{eq:dac_conv} can replace the usual ones (by
removing the post-activation of the layer before the one considered,
but see Remark~\ref{rm:pre_and_post_activation}).  In a few cases this
change could have no real effect: in fact there are situations in
which the standard shared biases are effectively unshared, notably
when the subsequent layer has only $n=1$ unit/kernel, or if it is a
depthwise convolution~\cite{ChoXDL}.  In these cases DAC reduces to a
standard connection.  In all other situations, DAC will use more
parameters and require more computations than a standard connection.

\textbf{Parameters and number of operations.}
A dense layer with $n$ units and $m$ inputs, has $mn$ weights.  When
it is pre-activated with unshared biases as in~\eqref{eq:dac_dense}, a
total of $mn$ DAC biases are used, in place of the $m$ shared biases
in the post-activation of the layer before that one, for a relative
increase factor $2-1/n$, that is, almost 2 when $n$ is not very small.
For a convolutional layer as in~\eqref{eq:dac_conv}, the number of
biases involved is the same, but the weights are $mnL^2$, so the
relative increase factor is $1+1/L^2-1/nL^2$. In the case of a
$3\times 3$ kernel, this leads to a modest increase of approximately
$+11\%$ in the number of parameters.

In calculating FLOPs, we adopt the usual convention to ignore
activation function costs.  Though pre-activation involves
significantly more calls to the activation function than
post-activation, for simple functions like ReLU, the computation cost
is very small and can be neglected.  This assumption may not hold for
other, more computationally expensive, activation functions.

Considering again a dense layer with $n$ units and $m$ inputs, FLOPs
are $mn$ multiplications and $mn$ additions.  When using
pre-activation, another $mn$ additions are required, that replace the
$m$ additions in the post-activation of the layer before that one.
The relative increase factor is $1.5-1/2n$, that is, almost $+50\%$
when $n$ is not very small.  In the case of a convolutional layer,
FLOPs per unit/kernel are $mL^2$ multiplications and $mL^2$ additions,
totaling $2mnL^2st$ operations for an output shape of $s\times
t$. When using pre-activation as in~\eqref{eq:dac_conv}, since DAC
biases $b_{i,j}$ do not depend on the particular kernel position,
initial activation results $\varphi(b_{i,j}+z_{\cdot,\cdot,j})$ can be
\emph{cached}, requiring $mnst$ additions, in place of the $mst$
additions in the post-activation of the layer before that one.  This
results in a relative increase factor of $1+0.5L^{-2}(1-1/n)$, or
about $+5.5\%$ for $3\times3$ convolutions and $+50\%$ for $1\times1$
convolutions.

\textbf{More complex structures.}  While the above discussion explains
how to use DACs in the case of a plain network, i.e.~a regular
sequence of basic layers, it is not always clear what to do in more
realistic situations, that include, for example, normalization layers
and skip connections.  The guiding principle then, is to identify
activations followed by linear operators, and check whether, from the
point of view of the linear operators, those activations are using
shared biases.  If this is the case, using DAC means to remove said
activations and add pre-activations with unshared biases to the linear
operators.

Hence for example, in a periodic sequence like
\[
\ldots\to\text{ReLU}\to\text{linear}\to\text{BN}\to\text{ReLU}\to\text{linear}\to\dots
\]
where BN stands for batch normalization and ``linear'' might be either
fully connected or convolutional layers, a DAC version would be
something like
\[
\ldots\to\text{DAC}\to\text{BN}'\to\text{DAC}\to\dots
\]
where $\text{BN}'$ is batch normalization without the trainable shift
parameter $\beta$ and DAC might be either~\eqref{eq:dac_dense}
or~\eqref{eq:dac_conv}.

\section{Experiments}
\label{sec:experiments}

We tested empirically the effect of using pre-activations with
unshared biases, on several common tasks.  All the experiments can be
reproduced using the source code~\cite{CurDAC}, that includes a test
implementation of the pre-activated layers in~\eqref{eq:dac_dense}
and~\eqref{eq:dac_conv}.

\textbf{SGEMM performance regression.}
We experimented with a regression task from the UCI
repository~\cite{sgemm2018,UCIKER}.  The task is to predict the
execution time of a matrix multiplication on a highly-tuneable SGEMM
kernel for GPU.  The input variables are 14 kernel parameters that are
either binary or take values that are powers of 2, and the response
variable ranges between 13.25 and 3397.08 milliseconds.  Non-binary
variables, including the response, were $\log_2$-transformed.  Input
variables were normalized and the dataset, consisting of all the
241600 combinations, one replica each, was split with a 70\%, 15\%
and 15\% proportion between training, validation and test.  We
assessed performances as mean squared error (MSE) between predicted
and real response (in $\log_2$ scale).

The nature of the task suggests to use fully connected neural
networks.  We tested several hundreds architectures, all with the same
basic structure but variable size.  The structure is a simple sequence
of fully connected layers, each followed by batch normalization and
ReLU.  The output is a fully connected layer with 1 unit and no
activation.  The size ranges in depth (from 5 to 16 layers before the
last one), in width (from 16 to 250 units) and in the progression of
widths with layers, that was either rectangular (constant width) or
pyramidal (decreasing width).  The number of trainable parameters
ranged between 1700 and 1.4M.

Network structures were defined in pairs, consisting of one network
with post-activations and shared biases (baseline), and one network
with pre-activations and unshared biases (DAC) with similar number of
trainable parameters.  To this end, in the DAC version, the number of
units in each layer was reduced by roughly a factor $\sqrt2$, in such
a way that the number of weights was halved, and adding the unshared
biases, the total number of trainable parameters was almost exactly
the same as for the baseline network.

For the training we used MSE loss, L$^2$-regularization of weights
(not biases) with a coefficient of $10^{-5}$, Adam optimizer with a
batch size of 1024, learning rate set to 0.01, He normal initializer for weights, and zero
initializer for biases (both shared and unshared).
Early stopping based on validation MSE
was activated after epoch 250, with a patience of 50 epochs and a
threshold of $10^{-4}$.

Training failed to reach a suitable error level in 15/576 cases for
the baseline and 2/449 cases for DAC.  The occurrence of this problem
is significantly lower for DAC (bilateral p-value 0.0063).  Failed
training experiments were excluded from further analysis as outliers.
Figure~\ref{fig:fc_all_experiments} in the Appendix shows the results of all the other
experiments.

\begin{figure}[tb]
  \centering
  \includegraphics[width=\columnwidth, trim=0 0 0 0, clip]{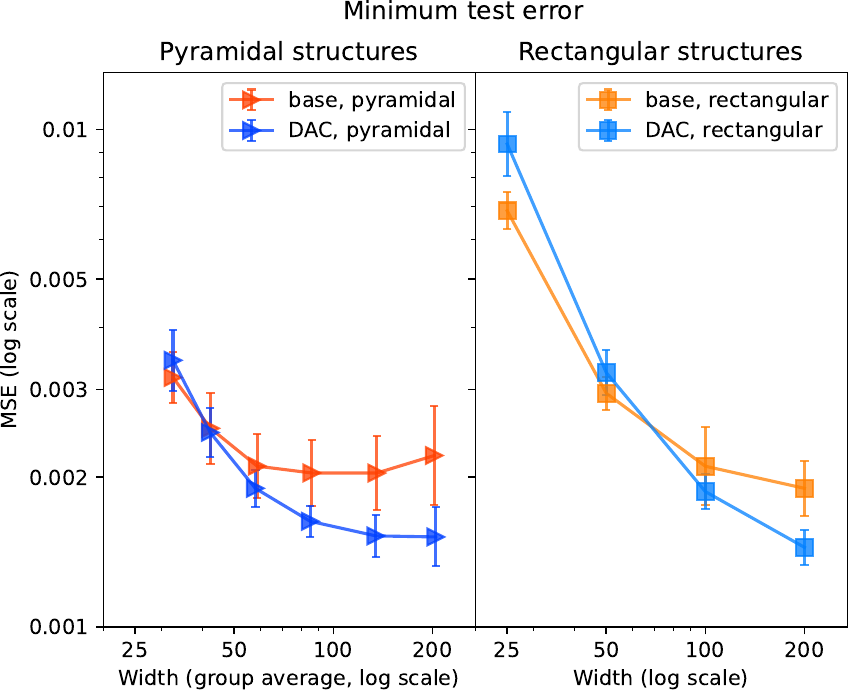}
  \caption{Aggregated and averaged results for the SGEMM regression
    task.  Experiments are grouped by network shape (pyramidal or
    rectangular, see text) and width.  Error bars represent the sample
    standard deviation of the values concurring to the average.
    Fully connected networks with DACs perform better than the
    baseline for larger widths, and similarly or worse for smaller
    widths, when the general performance of the network is far from
    optimal.}
  \label{fig:fc_progr_const}
\end{figure}
We found that there is weak dependence on depth for this task, and
hence we aggregated experiments by width, see
Figure~\ref{fig:fc_progr_const}.  Since DAC networks with the same
number of parameters have reduced number of units, we computed their
``equivalent width'', i.e.~$\sqrt2$ times the actual width: for example
rectangular DAC networks with 141 units per layer have as many
parameters as base networks with the same depth and 200 units per
layers, and so we say they have equivalent width 200.  For rectangular
networks, there are four groups with widths (or equivalent widths) 25,
50, 100 and 200 for base (or DAC).  For pyramidal networks, the width
of layers is not uniform, so we estimated an average width (or
equivalent width) as $\sqrt{\text{parameters}/\text{depth}}$.  Obtained
values formed six natural clusters that were used as groups.

\begin{figure}[bt]
  \centering
  \includegraphics[width=\columnwidth, trim=0 0 0 0, clip]{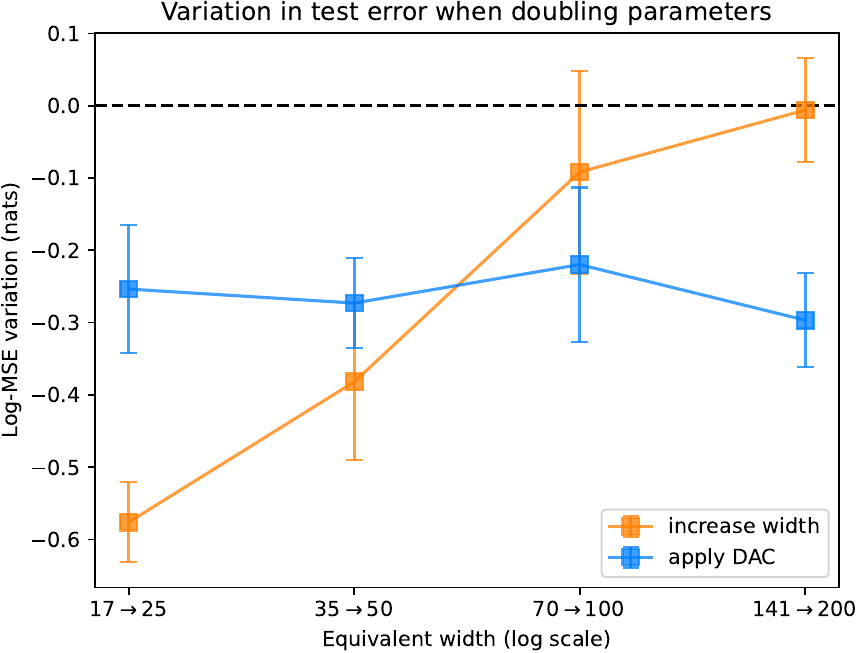}
  \caption{Efficiency analysis of unshared biases for the SGEMM
    regression task.  Rectangular baseline networks were compared with
    models with double the parameters: either by adding weights
    (orange) or by making biases unshared (blue).  The resulting
    variations of the MSE are shown (negative means improvement).
    Error bars represent the sample standard deviation of the values
    concurring to the average (see text).}
  \label{fig:fc_double_pars}
\end{figure}
From these initial analyses we observed that passing from shared to
unshared biases can be inefficient if the network is too small and
hence has insufficient capacity.  To better explore the matter of
efficiency of biases with respect to weights, we considered a
situation in which one wants to enlarge a standard (shared-biases)
model that has $N$ weights, either by adding weights or by adding
biases.  We measured the error reduction obtained when the number of
parameters doubles in these two ways: by increasing the number of
units by a factor $\sqrt2$ (adds about $N$ weights), or by making the
biases unshared (adds about $N$ biases).  To estimate this reduction,
first we obtained for each experiment the averages of log-MSE across
replicates; then we computed the differences between models of size
normal and double, for all widths and depths; then, since these
differences did not show dependence on the depth (see
Figure~\ref{fig:doubling_depth_analysis} in the Appendix), for each
width we collected the different depths and computed the average of
the values.

The results are shown in Figure~\ref{fig:fc_double_pars}.  Increasing
the number of weights (orange) has diminishing returns when the width
increases, with large benefits for small sizes and almost no
improvement passing from 141 to 200 units per layer.  Passing from
shared to unshared biases instead (blue), gives a uniform improvement
of about 0.25 nats in the test error.  This confirms that small
networks, with error levels that are far from optimal, benefit more
from increased width than from unshared biases, but when further width
increase is no more beneficial, then using unshared biases gives an
additional boost of performances and leads to otherwise unreacheable
error levels.

\textbf{Image classification tasks.}
We further experimented with convolutional architectures for image
classification tasks, like VGG and ResNet.  In all experiments we
trained a standard structure with post-activations and shared biases
(baseline) and a similar network with pre-activations and unshared
biases (DAC).  In the case of convolutional layers, using unshared
biases only increases marginally the number of parameters and the
number of operations (see Section~\ref{sec:methods}), so we did not
reduce the number of units in the DAC version.

In all experiments we trained with the following hyperparameters:
cross-entropy loss; L$^2$-regularization of kernel weights, with
coefficient tuned for the baseline and left the same for DAC; no
regularization of biases; initialization zero for all biases and He
normal for weights; data augmentation as proposed, for CIFAR,
in~\cite{HZRDRL}; batch size 128; SGD optimizer with momentum 0.9;
learning rate schedule starting at 0.1 and decreasing by a factor 10
after 40\%, 60\% and 80\% of the total training steps, with the total
length depending on the dataset size.

We chose three classification tasks with datasets of diverse nature:
CIFAR-10 and CIFAR-100~\cite{KNHCIF}; two subsets of ImageNet called
Imagenette and Imagewoof~\cite{HowIMA}; and ISIC 2019, a medical
images dataset, consisting of skin lesion images~\cite{ISIC19}.  Below
we give further details and results of the various experiments.

\begin{figure}[htb]
        \centering
        \includegraphics[width=1\columnwidth, trim=0 0 0 0, clip]{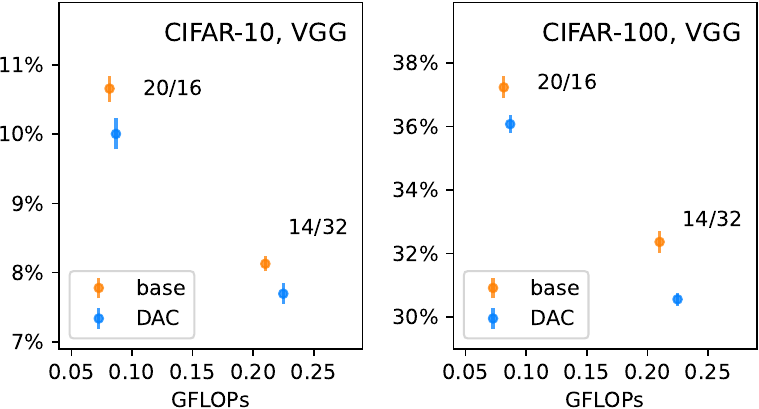}
        \caption{VGG, average test error. Compared performances of VGG
          20 layers, 16 channels, and VGG 14 layers, 32 channels with
          shared biases (baseline, orange) and unshared biases (DAC,
          blue) on CIFAR-10 and CIFAR-100. Test error (vertical axis)
          is averaged over 5 replicates and over 5 epochs (see
          text). Error bars are 95\% confidence intervals for the true
          mean value. Complexity (horizontal axis) is measured in
          GFLOPs per forward pass.}
        \label{fig:vgg}
\end{figure}

\textbf{Plain convolutional networks on CIFAR.}  We designed two
VGG-like architectures for CIFAR-10 and CIFAR-100, in line with modern
revisitations like~\cite{ding_repvgg_2021}.  The structure is a
simple sequence of $3\times3$ convolutional layers, each followed by
batch normalization and ReLU.  The output is a global average pooling
(GAP) layer, followed by a fully connected layer with 10 or 100 units
and softmax activation.  The first structure (referred as 20/16) has
20 layers including the output, and the 19 convolutional layers start
with 16 kernels that become 32 and 64, after 7 and 13 layers.  The
second structure (14/32) has 14 layers, starts with 32 kernels and
doubles after 5 and 9.

The split ratio between training, validation and test was set to
4:1:1.  We trained for a total of 80k steps, corresponding to 256
epochs.  The coefficient of L$^2$ regularization was $2\cdot10^{-4}$
for CIFAR-10 and $3\cdot10^{-4}$ for CIFAR-100.  Each experiment was
replicated 5 times with a 5-fold cross-validation scheme.  The best
test accuracy was estimated by averaging over the 5 replicates and
over 5 epochs centered on the best epoch on the validation dataset,
see Section~\ref{sec:Error rate estimation} of the Appendix for
details on this robust statistical estimator.

The networks with pre-activations and unshared biases obtained a
systematic improvement in the test accuracy with respect to the
baselines.  The improvement is statistically significant in all
cases.  For the two architectures 20/16 and 14/32, the improvement was
$0.65\%\pm0.17\%$ and $0.43\%\pm0.09\%$ on CIFAR-10, and it was
$1.16\%\pm0.22\%$ and $1.81\%\pm0.19\%$ on CIFAR-100, see
Figure~\ref{fig:vgg}.  The models with unshared biases require only a
marginal increase in complexity, with 11\% more parameters and 5.5\%
more FLOPs than the baseline, so these improvements cannot be
explained just by the larger size of the models.

\begin{remark}
To measure the size of the models, in this and other plots, we favor
floating point operations (FLOP) over the number of parameters, as was
advocated among other sources in~\cite{SDSGAI}.  We provide plots with
the number of parameters in the Appendix.
\end{remark}

\textbf{ResNet networks on CIFAR.} We used as baseline the
architecture proposed for CIFAR in the original ResNet
papers~\cite{HZRDRL, HZRIMD}, with 20, 32, 44, and 56 layers ($n=3, 5,
7, 9$) both with the v1 post-activated~\cite{HZRDRL} architecture and
the v2 pre-activated~\cite{HZRIMD} architecture.  The structure is
similar to the VGG of the previous section, but with skip connections
every two layers.  More explicitly, it starts with a convolution with
16 kernels, followed by three stages of $n$ residual blocks each, with
16, 32 and 64 kernels respectively.  The output is GAP plus fully
connected, as before.

The residual blocks come in two versions, referred as v1 and v2.  For
v1 there is a classical post-activation sequence:
\[
\text{conv}\to\text{BN}\to\text{ReLU}\to\text{conv}\to\text{BN}\to+\text{input}\to\text{ReLU}
\]
For v2, instead, the authors found that ResNet performs better with
a pre-activations sequence:
\[
\text{BN}\to\text{ReLU}\to\text{conv}\to\text{BN}\to\text{ReLU}\to\text{conv}\to+\text{input}
\]

The conversion to pre-activation with unshared biases (DAC) was done
as follows:
\begin{gather*}
  \text{DAC}\to\text{BN}\to\text{DAC}\to\text{BN}\to+\text{input}
  \tag{v1}\\
  \text{BN}\to\text{DAC}\to\text{BN}\to\text{DAC}\to+\text{input}
  \tag{v2}
\end{gather*}

\begin{figure}[tb]
        \centering
        \includegraphics[width=\columnwidth, trim=0 0 0 0, clip]{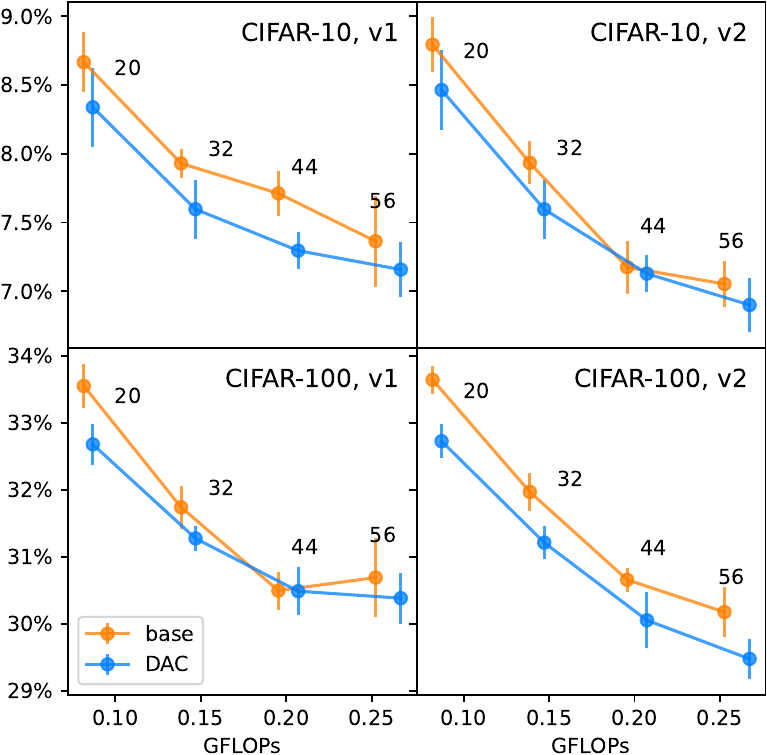}
        \caption{ResNet, average test error.  Compared performances of
          ResNet networks with shared biases (baseline, orange) and
          unshared biases (DAC, blue), on CIFAR-10 (top) and CIFAR-100
          (bottom), with architectures v1 (left) and v2 (right).
          Floating numbers are the layers count.  Test error (vertical
          axis) is averaged over 5 replicates and over 5 epochs (see
          text).  Error bars are 95\% confidence intervals for the
          true mean value.  Complexity (horizontal axis) is measured
          in GFLOPs per forward pass.}
        \label{fig:comparison_results}
\end{figure}
The training hyperparameters were the same as in the plain
convolutional networks experiments.

Figure~\ref{fig:comparison_results} summarizes the results in terms of
best test accuracy, estimated robustly as for the VGG experiments (see
Section~\ref{sec:Error rate estimation} of the Appendix).  Among the
16 comparisons, 15 are in favor of DAC models, with 9 of them
statistically significant (p-value below 5\%).  This confirms that
using unshared biases improves the performances of residual
convolutional networks on CIFAR, with only a marginal increase in the
model complexity.  In fact, one could argue that, if the improvement
of the test error between baseline and DAC was due only to the larger
sizes of the latter, then in the figure, the decreasing blue and
orange lines connecting models of growing depth, would be
superimposed.  They are partially superimposed for CIFAR-10, v2 and
CIFAR-100, v1, but well separated in the other two cases.

\begin{table}[tb]
\caption{Comparison of minimum test errors in percent points on CIFAR-10 and CIFAR-100 ($\pm$ one standard error).}
\label{fig:table}
\centering
\begin{small}
\begin{tabular}{c@{\hspace{8pt}}c@{\hspace{5pt}}c@{\hspace{8pt}}c@{\hspace{5pt}}c}
\toprule
{}   & \multicolumn{2}{c}{CIFAR-10, v1}                     & \multicolumn{2}{c}{CIFAR-10, v2}                                     \\ \midrule 
{Layers} & {Base}   & {DAC}             & {Base}   & {DAC}             \\  
{20} & {8.64$\pm$0.12} & {\textbf{8.28}$\pm$0.19} & {8.73$\pm$0.10} & {\textbf{8.33}$\pm$0.07} \\
{32} & {7.90$\pm$0.04} & {\textbf{7.57}$\pm$0.11} & {7.85$\pm$0.09} & {\textbf{7.56}$\pm$0.15} \\
{44} & {7.69$\pm$0.09} & {\textbf{7.27}$\pm$0.06} & {7.14$\pm$0.09} & {\textbf{7.10}$\pm$0.05} \\
{56} & {7.34$\pm$0.17} & {\textbf{7.13}$\pm$0.11} & {7.02$\pm$0.09} & {\textbf{6.89}$\pm$0.05} \\
\midrule
{}   & \multicolumn{2}{c}{CIFAR-100, v1}                    & \multicolumn{2}{c}{CIFAR-100, v2}                                                                     \\  
\midrule
{Layers}     & {Base}             & {DAC}              & {Base}    & {DAC}  \\  
{20} & {33.50$\pm$0.16} & {\textbf{32.54}$\pm$0.04} & {33.50$\pm$0.11} & {\textbf{32.62}$\pm$0.10} \\
{32} & {31.66$\pm$0.12} & {\textbf{31.09}$\pm$0.09} & {31.79$\pm$0.12} & {\textbf{31.05}$\pm$0.12} \\
{44} & {\textbf{30.40}$\pm$0.11} & {30.42$\pm$0.19} & {30.50$\pm$0.07} & {\textbf{29.93}$\pm$0.20} \\
{56} & {30.53$\pm$0.23} & {\textbf{30.32}$\pm$0.19} & {29.89$\pm$0.22} & {\textbf{29.32}$\pm$0.17} \\
\bottomrule
\end{tabular}
\end{small}
\end{table}

Table~\ref{fig:table} presents a simpler metric: the lowest test error
rate (calculated as the minimum over the epochs of the average of the
5 replicates). The performance of the baselines aligns with typical
values for networks of similar complexity found in the literature.
Both the figure and table suggest that using unshared biases in ResNet
architectures leads to measurable performance improvements across most
versions and benchmark datasets. In some instances, using unshared
biases improves the corresponding baselines despite having fewer
layers.

We performed also an experiment on CIFAR-10 using a four times wider
ResNet20 v2 architecture, with 64, 128 and 256 kernels in the three
stages, resulting in a total of 4.3M parameters for the baseline
model.  Using unshared biases instead of shared biases, the minimum
test error dropped from 5.69\% to 5.16\%, while the number of
parameters increased as usual by 11\%, and the FLOPs by 5.5\%.

\begin{table}
\caption{Comparison of minimum test errors in percent points on Imagenette and Imagewoof ($\pm$ one standard error).}
\label{fig:table_imagenette} 
\centering
\begin{small}
\begin{tabular}{@{\hspace{4pt}}c@{\hspace{7pt}}c@{\hspace{4pt}}c@{\hspace{7pt}}c@{\hspace{4pt}}c@{\hspace{4pt}}} 
\toprule
{} & \multicolumn{2}{c}{Imagenette, v1} & \multicolumn{2}{c}{Imagenette, v2} \\
\midrule 
{Layers} & {Base} & {DAC} & {Base} & {DAC} \\
{20} & {13.42$\pm$0.26} & {\textbf{11.69}$\pm$0.11} & {11.98$\pm$0.17} & {\textbf{11.79}$\pm$0.13} \\
{32} & {13.14$\pm$0.45} & {\textbf{11.75}$\pm$0.08} & {11.40$\pm$0.09} & {\textbf{11.28}$\pm$0.11} \\
\midrule
{} & \multicolumn{2}{c}{Imagewoof, v1} & \multicolumn{2}{c}{Imagewoof, v2} \\
\midrule
{Layers} & {Base} & {DAC} & {Base} & {DAC} \\
{20} & {23.20$\pm$0.26} & {\textbf{22.61}$\pm$0.25} & {22.61$\pm$0.17} & {\textbf{21.70}$\pm$0.22} \\
{32} & {23.06$\pm$0.24} & {\textbf{21.32}$\pm$0.29} & {21.75$\pm$0.17} & {\textbf{20.65}$\pm$0.14} \\
\midrule
{\multirow{2}{*}{GFLOPs}} & {0.509} & {0.542} & {0.509} & {0.542} \\
{} & {0.865} & {0.918} & {0.865} & {0.918} \\
\bottomrule
\end{tabular}
\end{small}
\end{table}

\textbf{ResNet networks on Imagenette and Imagewoof.}  Imagenette and
Imagewoof~\cite{HowIMA} are subsets of ImageNet that are frequently
utilized for model benchmarking. They offer a simpler and faster
alternative to ImageNet while preserving many of its
challenges. Imagenette comprises about 10k images belonging to 10 easily
distinguishable classes, whereas Imagewoof includes 10 classes that
are more challenging to classify due to their similarities, as they
represent 10 different dog breeds.

For each dataset, we selected the 160-pixel version, in which the
shortest side is resized to 160 pixels while maintaining the aspect
ratio.  These images undergo further processing to achieve a final
size of $80\times80$ pixels.
The split ratio between training and validation was set to 4:1:2
approximately.  We trained for a total of 64k steps, corresponding to
259 epochs.

We used the 20 and 32 layers ResNet structure of the CIFAR-10 experiments, with the same training hyperparameters.

Table~\ref{fig:table_imagenette} summarizes the resulting lowest test
error rates (calculated as the minimum over the epochs of the average
of the 5 replicates).  The performance of the baselines aligns with
typical values for networks of similar complexity found in the
literature.  Using unshared biases (DAC) shows marked performance
improvements, again at the cost of a marginal increase in size and
complexity.  In particular it can be observed that the 
baseline ResNet v1 (post-activated) is much worse than baseline v2 and DACs (both pre-activated).

\textbf{ResNet networks on images for melanoma diagnosis.}  To investigate
performances in image classification tasks that are both more
difficult and more applied, we conducted an experiment using a ResNet
v1 model on a real-world dataset from \emph{International Skin Imaging
Collaboration} (ISIC).  We used ISIC 2019~\cite{ISIC19}, a collection of 25330
quality-controlled dermoscopic images of skin lesions, divided into 8
classes.  With 20 layers, the baseline error is 27.05\%, and with DAC
is 26.47\%, for an improvement of 0.58\%.  The same figures for 32
layers are 26.30\% and 25.04\%, for an improvement of 1.26\%.

\begin{table}
\caption{Ablation study isolating the contribution made by
  pre-activation only (with shared biases) for ResNet20 architectures
  on CIFAR.  Reported values (absolute) are the estimated improvements
  in test accuracy when replacing the standard post-activation with
  pre-activation.  Relative values are relative with respect to the
  estimated improvement when using full DACs, with pre-activations
  and unshared biases.
}
\label{fig:ablation}
\begin{center}
  \begin{small}
  \begin{tabular}{lcccc}
    \toprule
              & \multicolumn{2}{c}{Absolute}
                                      & \multicolumn{2}{c}{Relative to DAC} \\
              &        v1 &        v2 &      v1 &      v2 \\
    \midrule
    CIFAR-10  & $-0.13\%$ & $+0.08\%$ & $-41\%$ & $+31\%$ \\
    CIFAR-100 & $+0.08\%$ & $+0.07\%$ &  $+9\%$ &  $+8\%$ \\
    \bottomrule
  \end{tabular}
  \end{small}
\end{center}
\end{table}

\textbf{Ablation study: pre-activation with shared biases.}  Since
unshared biases cannot be realized without pre-activations, in all
previous DAC experiments we used both together.  We decided then to
investigate briefly with pre-activation only (hence, with shared
biases).  We trained ResNet20 v1 and v2 architectures, on CIFAR-10 and
CIFAR-100.

We found that these models performed comparably to the baseline model,
but distinctly worse than using full DACs, with pre-activations and
unshared biases, see Table~\ref{fig:ablation}.  In one case the result
was slightly worse than the baseline, and even in the other cases the
improvements were never statistically significant.

\textbf{Comparison with MBA.}  Since multi-bias activation (MBA, see
Section~\ref{sec:related}) introduced in~\cite{LOWMBA} is based on a
principle close to DAC, we tested it on the same ResNet20 architecture
on CIFAR-100.  MBA works by creating $K$ copies of the output of
convolutions, then applying independently trainable biases to each of
them, applying ReLU, and then using the enlarged output as input for
the subsequent layer.  When converting a baseline structure to MBA,
the number of parameters increases $K$-fold, so these models are much
larger than DAC, that increases the parameters by 11\%.  In spite of this, we found
improvements over the baseline smaller than using DACs (0.52\%,
0.64\%, 0.14\% for MBA with $K=2$, $4$, $8$, and 0.97\% for DAC).  It
is possible that MBA might require a specific tuning of
hyperparameters to reach better performances, but since we did not
do this for DAC models, we did not investigate the matter further.

\section{Theoretical discussion}
\label{sec:theoretical_main}

In Section~\ref{sec:experiments} we gathered empirical evidence
supporting the primary assertion of this paper: the use of
pre-activations with unshared biases, trading some weights for
additional biases, can be an efficient way to boost a model's
performance.  This section explores theoretical arguments that support
this claim, highlighting the fact that using unshared biases improves
considerably the flexibility of the model.

\textbf{Last and first layers.}  In a plain fully connected neural
network with ReLU activations and standard units, one would expect a
ReLU activation at the output of the very last layer.  However, this
is undesirable as the final output should be informative and
compatible with the loss function (e.g., logits for cross-entropy).
So one usually has to remove that last activation, which, using
pre-activations, would not have existed in the first place.

Symmetrically, consider the first layer of a similar network: with
standard units, it would not make sense to filter the input nodes with
ReLU.  Nevertheless, with unshared biases, it is instead very
reasonable to apply the nonlinearity to the input nodes, because
different units in the first layer might benefit from tailored
filtering of the input.

In both cases pre-activated units with unshared biases seem more
natural.

\textbf{Input replication.}  Filtering the input as just described
might be even more useful if the input is replicated multiple times.
Consider the toy example of a one-dimensional input $x$ and a shallow
network with only one layer of one unit aiming at approximating some
function $f:\R\rightarrow\R$.  A pre-activated unit `$0$' with input
replicated $n$ times $\overline x=(x,x,\dots,x)$ and unshared biases
gives:
\begin{equation}
\label{eq:replication}
\hat{f}_{\text{DAC}}(\overline x)
=\sum_{j=1}^n w_{0,j}\, \varphi(b_{0,j}+x),
\end{equation}
which is a universal approximator of a large class of functions
$\R\rightarrow\R$, for $n\to\infty$. On the other hand, a standard
post-activated unit with replicated inputs would give:
\[
\hat{f}_{\text{std}}(\overline x)
=\varphi\biggl(b_{0} + \sum_{j=1}^n w_{0,j}\, x\biggr)
=\varphi(b_{0} + \tilde w_{0}\, x),
\]
equivalent to the same without replicating the inputs, regardless of
$n$.  To gain expressivity we can add a hidden layer with $n$ standard
units (with or without replicated inputs is the same), obtaining:
\[
\hat{f}_{2\times\text{std}}(\overline x)
=\varphi\biggl(b_0+\sum_{j=1}^n w_{0,j}\, \varphi(b_j+\tilde{w}_{j}\, x)\biggr).
\]
To show that $\hat{f}_{2\times\text{std}}$ has a representation power
similar to $\hat{f}_{\text{DAC}}$, we reparametrize, putting
$\tilde{w}_{0,j}=w_{0,j}|\tilde{w}_{j}|$ and
$\tilde{b}_j=\nicefrac{b_j}{|\tilde{w}_{j}|}$ in the above expression,
which gives:
\[
\hat{f}_{2\times\text{std}}(\overline x)
=\varphi\biggl(b_0+\sum_{j=1}^n \tilde w_{0,j}\, \varphi\bigl(\tilde b_j+\sign(\tilde{w}_{j})\, x\bigr)\biggr).
\]
Thus, in this toy problem, one needs a two-layers standard neural network to
get a representation power similar to a single pre-activated unit.


\textbf{Backpropagation.}  In this section, we derive the
backpropagation equations for a fully connected multi-layer plain
architecture with pre-activations using unshared biases.
Interestingly, these models enable a more granular masking of the
various contributions to the gradient.

We compute the derivatives of a loss $E$ with
respect to the parameters.  Let $y_i$, $w_{i,j}$ and $b_{i,j}$ denote
the outputs, weights, and DAC biases of layer $k$, respectively, and
let $m$ denote the number of units and $y^\diamond_j$ denote the
outputs of layer $k-1$.  Then
$y_i=\sum_{j=1}^m w_{i,j}\,\varphi(b_{i,j}+y^\diamond_j)$, and we get:
\begin{equation}
\label{eq:backpropagation}
\begin{cases}
\frac{\partial E}{\partial w_{i,j}}
=\varphi(b_{i,j}+y^\diamond_j)\,
\frac{\partial E}{\partial y_{i}} \\
\frac{\partial E}{\partial b_{i,j}}
=w_{i,j}\,\varphi'(b_{i,j}+y^\diamond_j)\,
\frac{\partial E}{\partial y_{i}}\\
\frac{\partial E}{\partial y_{i}} 
=\sum_{l=1}^n
w^*_{l,i}\,\varphi'(b^*_{l,i}+y_i)
\frac{\partial E}{\partial y^*_{l}}
\end{cases}
\end{equation}
Here $n$, $w^*_{l,i}$, $b^*_{l,i}$ and $y^*_l$ are the
units, the weights, the pre-activation biases and the outputs of layer
$k+1$, respectively.

If we were to use post-activated units with shared biases instead,
then $\varphi'(b^*_i+y_i)$ in the last equation would not depend on
$l$ and could be taken out of the sum.  This would result in a single
0-1 factor regulating the entire derivative.  Therefore, unshared
biases provide a more granular masking of the different contributions
to the gradient.

\section{Conclusions}
\label{sec:conclusions}

This paper is intended as a foundational study on architectures
that leverage pre-activation with unshared biases.  It provides
qualitative arguments and empirical evidence that this choice has measurable advantages and promising
potential and that there are situations in which trading some
weights for additional biases is more efficient. As a future development, one could investigate if and how DAC units can be integrated on diverse architectures, like for instance mobileNet \cite{SHZMIR}, EfficientNet \cite{TaLERM},  Transformers \cite{VSPAIA}, ViT \cite{DBKIWW}, generative models \cite{KiWAVB, GPMGAN} or in a reinforcement learning setting like the ones described in \cite{SHMMGG, WuDASP, MAGSSA, MAFSSAec, PPMSWS, MAG9x9}. 
See the last paragraph of Section~\ref{sec:methods} about DAC integration into complex models.


\paragraph{Acknowledgements.}
Many thanks to Rosa Gini for her important intellectual contribution.

\newpage

{\small
\bibliographystyle{ieee_fullname}
\bibliography{mau.bib}
}

\newpage
\appendix






\onecolumn
\begin{center}
    \Huge\textbf{Appendix}
\end{center}


\section{Biological inspiration}
\label{sec:bio_inspiration}

The proposed extension of the artificial neuron also reflects to some extent a recent shift in the understanding of the biological neuron.  In fact, the early soma-centric representation of the neuron today has been discarded in favor of a more realistic and complex model that incorporates \emph{active} dendrites~\cite{LarADC,SiNADL}.

A typical biological neuron consists of many input branches called dendrites, a main body called soma, and the axon, which branches at its end in many terminals, where synapses connect to the dendrites of other neurons.  The input signals originate in the dendrites, flow through the soma, and are integrated into the region of the soma where the axon connects, and if a specific threshold is reached, the neuron fires its signal down the axon, to the synapses.

Until some years ago, the biological neuron model was soma-centric and essentially modeled by a point neuron where dendrites simply pass the signals, and all elaboration happens at the soma.  This elementary representation was the inspiration of the traditional perceptron in artificial neural networks.

Current biological models are more complicated and the central role of dendrites in signal modulation is better understood~\cite{PoPIDF}.  Dendrites in fact present voltage-gated ion channels~\cite{LarADC} able to produce local electrical events termed dendritic spikes.  Dendrites actually present at least four groups of ion channels~\cite{SiNADL}: the synaptic receptors, activated by neurotransmitters, the passive leak channels, and the active subthreshold ion channels, able to produce transmembrane currents also when the threshold for the action potential is not reached, and supra-threshold ion channel active when the threshold is reached.
In this way, a dendrite or a group of dendrites can perform the first important local, not linear signal integration before reaching the cell axon.

\section{Error rate estimation}
\label{sec:Error rate estimation}

While we always trained for the same number of steps, we evaluated the best error rate by simulating early stopping.  Let $v_{k,j}$ and $t_{k,j}$ denote the validation and test errors for replicate $k$ and epoch $j$.  We select the epoch $m$ corresponding to the minimum validation error (averaging on the 5 replicates and on a moving window of 5 epochs), and then compute the average test error of the same 5 epochs and replicates:
\[
m:=\argmin_i\frac1{25}\sum_{k=1}^5\sum_{j=-2}^2v_{k,i+j}
,\qquad
\overline T:=\frac1{25}\sum_{k=1}^5\sum_{j=-2}^2t_{k,m+j}.
\]
This approach allows for evaluation of the statistical error of the estimator and it is more robust and reliable than simply taking the minimum of the test error, as in a real application one would be able to choose the early stopping on the validation set, but not on the test set.

Assuming that $t_{k,m+j}=\mu_m+\sigma\,Z_k+\tau\,Z_{k,j}$, with $\mu_m$ the true value, $Z_k$ and $Z_{k,j}$ independent standard Gaussian noises, and $\sigma$, $\tau$ coefficients measuring randomness in replicates and epochs, the square of the standard error of $\overline T$ is $\Var(\overline T)=\frac15\sigma^2+\frac1{25}\tau^2$. Here the two terms were conservatively estimated using respectively:
\[
\sigma^2\leq\sigma^2+\tau^2
\approx\frac14\sum_{k=1}^5(t_{k,m}-\overline T)^2
\qquad\text{and}\qquad
\frac15\tau^2
\approx\frac14\sum_{j=-2}^2(t_{*,m+j}-\overline T)^2.
\]

\section{Representation power}
\label{sec:representation_power}

\subsection{Separation of sets in low dimension}
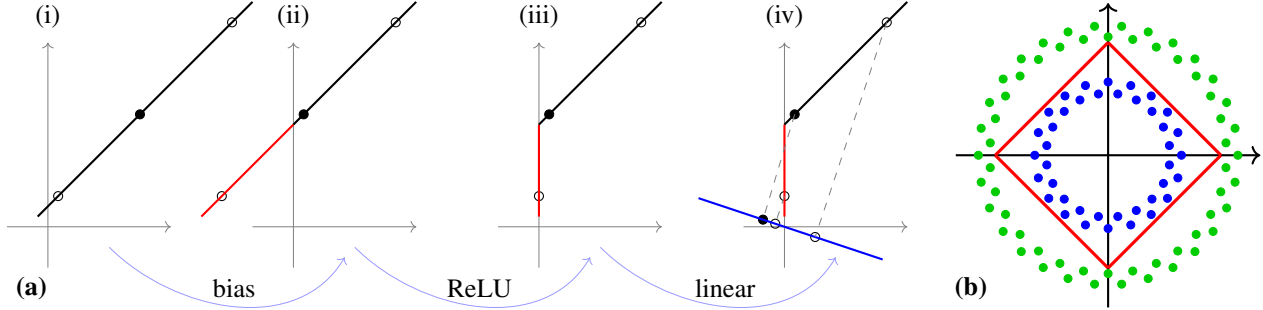
\begin{figure*}[tb]
\centering
\begin{subfigure}[ht]{0.66\textwidth}
\centering
\begin{tikzpicture}[scale=0.85]
    \def\step{0.32}

    \path[use as bounding box] (-26*\step,-19*\step) -- (18*\step,-4*\step);
    
    \path (-24*\step, -15*\step) coordinate (center);
    \path (center) + (-2*\step,0) coordinate (startx);
    \path (center) + (6*\step,0) coordinate (endx);
    \path (center) + (0,-2*\step) coordinate (starty);
    \path (center) + (0,9*\step) coordinate (endy);
    \path (center) +  (0.5*\step,1.5*\step) coordinate (leftpoint) ;
    \path (center) +  (4.5*\step,5.5*\step) coordinate (centerpoint) ;
    \path (center) +  (9*\step,10*\step) coordinate (rightpoint) ;
    \path (center) +  (-0.5*\step,0.5*\step) coordinate (startline) ;
    \path (center) +  (10*\step,11*\step) coordinate (endline) ;

    \path [draw, ->, very thin, gray] (startx) --  (endx);
    \path [draw, ->, very thin, gray] (starty) --  (endy);
    \draw (leftpoint) circle (2pt) ;
    \draw (rightpoint) circle (2pt) ;
    \filldraw[black] (centerpoint) circle (2pt);
    \path [draw, thick] (startline) --  (endline);

    \node[yshift = 1em] at (endy) {(i)} ;

    \path (-12*\step, -15*\step) coordinate (center);
    \path (center) + (-2*\step,0) coordinate (startx);
    \path (center) + (6*\step,0) coordinate (endx);
    \path (center) + (0,-2*\step) coordinate (starty);
    \path (center) + (0,9*\step) coordinate (endy);
    \path (center) +  (-3.5*\step,1.5*\step) coordinate (leftpoint) ;
    \path (center) +  (0.5*\step,5.5*\step) coordinate (centerpoint) ;
    \path (center) +  (5*\step,10*\step) coordinate (rightpoint) ;
    \path (center) +  (-4.5*\step,0.5*\step) coordinate (startline) ;
    \path (center) +  (0,5*\step) coordinate (centerline) ;
    \path (center) +  (6*\step,11*\step) coordinate (endline) ;

    \path [draw, ->, very thin, gray] (startx) --  (endx);
    \path [draw, ->, very thin, gray] (starty) --  (endy);
    \draw (leftpoint) circle (2pt) ;
    \draw (rightpoint) circle (2pt) ;
    \filldraw[black] (centerpoint) circle (2pt);
    \path [draw, thick, red] (startline) --  (centerline);
    \path [draw, thick] (centerline) --  (endline);

    \node[yshift = 1em] at (endy) {(ii)} ;

    \path (0, -15*\step) coordinate (center);
    \path (center) + (-2*\step,0) coordinate (startx);
    \path (center) + (6*\step,0) coordinate (endx);
    \path (center) + (0,-2*\step) coordinate (starty);
    \path (center) + (0,9*\step) coordinate (endy);
    \path (center) +  (0*\step,1.5*\step) coordinate (leftpoint) ;
    \path (center) +  (0.5*\step,5.5*\step) coordinate (centerpoint) ;
    \path (center) +  (5*\step,10*\step) coordinate (rightpoint) ;
    \path (center) +  (0*\step,0.5*\step) coordinate (startline) ;
    \path (center) +  (0,5*\step) coordinate (centerline) ;
    \path (center) +  (6*\step,11*\step) coordinate (endline) ;

    \path [draw, ->, very thin, gray] (startx) --  (endx);
    \path [draw, ->, very thin, gray] (starty) --  (endy);
    \draw (leftpoint) circle (2pt) ;
    \draw (rightpoint) circle (2pt) ;
    \filldraw[black] (centerpoint) circle (2pt);
    \path [draw, thick, red] (startline) --  (centerline);
    \path [draw, thick] (centerline) --  (endline);

    \node[yshift = 1em] at (endy) {(iii)} ;

    \path (12*\step, -15*\step) coordinate (center);
    \path (center) + (-2*\step,0) coordinate (startx);
    \path (center) + (6*\step,0) coordinate (endx);
    \path (center) + (0,-2*\step) coordinate (starty);
    \path (center) + (0,9*\step) coordinate (endy);
    \path (center) +  (0*\step,1.5*\step) coordinate (leftpoint) ;
    \path (center) +  (0.5*\step,5.5*\step) coordinate (centerpoint) ;
    \path (center) +  (5*\step,10*\step) coordinate (rightpoint) ;
    \path (center) +  (0*\step,0.5*\step) coordinate (startline) ;
    \path (center) +  (0,5*\step) coordinate (centerline) ;
    \path (center) +  (6*\step,11*\step) coordinate (endline) ;
    \path (center) +  (-0.45*\step,0.15*\step) coordinate (leftproj) ;
    \path (center) +  (1.5*\step,-0.5*\step) coordinate (rightproj) ;
    \path (center) +  (-1.05*\step,0.35*\step) coordinate (centerproj) ;
    \path (center) +  (-4.2*\step,1.4*\step) coordinate (startproj) ;
    \path (center) +  (4.8*\step,-1.6*\step) coordinate (endproj) ;

    \path [draw, ->, very thin, gray] (startx) --  (endx);
    \path [draw, ->, very thin, gray] (starty) --  (endy);
    \draw (leftpoint) circle (2pt) ;
    \draw (rightpoint) circle (2pt) ;
    \filldraw[black] (centerpoint) circle (2pt);
    \path [draw, thick, red] (startline) --  (centerline);
    \path [draw, thick] (centerline) --  (endline);
    \draw (leftproj) circle (2pt) ;
    \draw (rightproj) circle (2pt) ;
    \filldraw[black] (centerproj) circle (2pt) ;
    \path [draw, thick, blue] (startproj) --  (endproj);
    \path [draw, very thin, gray, dashed] (leftpoint) --  (leftproj);
    \path [draw, very thin, gray, dashed] (centerpoint) --  (centerproj);
    \path [draw, very thin, gray, dashed] (rightpoint) --  (rightproj);

    \node[yshift = 1em] at (endy) {(iv)} ;

    \path (-21*\step,-16*\step) coordinate (labela) ;
    \path (-9.5*\step,-16.5*\step) coordinate (labelaend) ;
    \path (-9*\step,-16*\step) coordinate (labelb) ;
    \path (2.5*\step,-16.5*\step) coordinate (labelbend) ;
    \path (3*\step,-16*\step) coordinate (labelc) ;
    \path (15*\step,-16*\step) coordinate (labeld) ;
    \path (14.5*\step,-16.5*\step) coordinate (labeldend) ;
    \path[->] (labela) edge[very thin, blue!40!white, in = -120, out = -45] node[black, above] {bias} (labelaend) ;
    \path[->] (labelb) edge[very thin, blue!40!white, in = -120, out = -45] node[black, above] {ReLU} (labelbend) ;
    \path[->] (labelc) edge[very thin, blue!40!white, in = -120, out = -45] node[black, above] {linear} (labeldend) ;
    
    \path (-24.8*\step,-18*\step) node {\textbf{(a)}};
\end{tikzpicture}
\end{subfigure}
\hfill
\begin{subfigure}[ht]{0.33\textwidth}
\centering
\begin{tikzpicture}[scale=0.75]
    \tikzstyle{formatcircle} = [ circle, radius =.2pt]
    \def\step{2.7}
    \def\steporangein{2.1}
    \def\steporangeout{2.3}
    \def\numcirc{30}
    \def\stepbluein{1.1}
    \def\stepblueout{1.3}
    \def\numcircblue{20}
    \def\stepred{2.0}

    \path (1.1*\step, -\step) coordinate (center);

    \path (center) + (-\step,0) coordinate (startx);
    \path (center) + (\step,0) coordinate (endx);
    \path (center) + (0,-\step) coordinate (starty);
    \path (center) + (0,\step) coordinate (endy);
    \path [draw, ->, thick] (startx) --  (endx);
    \path [draw, ->, thick] (starty) --  (endy);

    \foreach \n in {1,...,\numcirc} \path (center) +  (\n*360/\numcirc:\steporangeout) coordinate (Oout\n) ;
    \foreach \n in {1,...,\numcirc} \path (center) +  (.5*360/\numcirc + \n*360/\numcirc:\steporangein) coordinate (Oin\n);
    \foreach \n in {1,...,\numcirc} \filldraw[green!80!black] (Oout\n) circle (2pt) ;
    \foreach \n in {1,...,\numcirc} \filldraw[green!80!black] (Oin\n) circle (2pt) ;

    \foreach \n in {1,...,\numcircblue} \path (center) +  (\n*360/\numcircblue:\stepblueout) coordinate (Oout\n) ;
    \foreach \n in {1,...,\numcircblue} \path (center) +  (.5*360/\numcircblue + \n*360/\numcircblue:\stepbluein) coordinate (Oin\n);
    \foreach \n in {1,...,\numcircblue} \filldraw[blue] (Oout\n) circle (2pt) ;
    \foreach \n in {1,...,\numcircblue} \filldraw[blue] (Oin\n) circle (2pt) ;

    \path (center) + (-\stepred,0) coordinate (W);
    \path (center) + (\stepred,0) coordinate (E);
    \path (center) + (0,\stepred) coordinate (N);
    \path (center) + (0,-\stepred) coordinate (S);
    \foreach \coordA/\coordB in {N/E,S/E,N/W,S/W}  \path [draw, -, very thick, color = red]  (\coordA) -- (\coordB) ;

    \path (0.2*\step,-1.873*\step) node {\textbf{(b)}};
\end{tikzpicture}
\end{subfigure}
\caption{\textbf{(a)}.  A pre-activated unit with unshared biases and
  two inputs can separate points that are not linearly separable.  (i)
  Black and white points are not linearly separable.  (ii) Applying
  two independent biases allows to choose a translation that moves the
  leftmost point to the second quadrant (red line) while keeping the
  other in the first one.  (iii) ReLU projects all points in the
  second quadrant onto the vertical axis (red line), making the
  dataset linearly separable.  (iv) The linear part of the unit learns
  a direction (blue line) onto which to project to separate the points.
  \textbf{(b)}. The combination of a pre-activated layer
  $f:\R^2\rightarrow\R^2$ with two units and a linear unit
  $g:\R^2\rightarrow\R$ can separate the blue from the green points.
  In fact, let $[f(x)]_i=\sum_{j=1,2}\,w_{ij}\varphi(b_{ij}+x_j)$ with
  $\varphi$ denoting ReLU, then it is enough to set $w_{ij}=1$,
  $b_{1j}=1$, $b_{2j}=0$ and $g(y)=y_1-2y_2-1$ to get
  $g(f(x))=\varphi(1+x_1)+\varphi(1+x_2)-2\varphi(x_1)-2\varphi(x_2)-1$,
  which is positive and equals $1-|x_1|-|x_2|$ inside the red square
  and is negative outside.  On the contrary, it is easy to see that if
  $f:\R^2\rightarrow\R^2$ was a standard fully connected layer, then
  for all choices of $f$ and $g$ the set where $g(f(x))\geq0$ would
  always be unbounded or empty.  A post-activated network requires
  four units in the first layer to achieve the same result.}
\label{fig:separable}
\end{figure*}

Figure~\ref{fig:separable}
presents two toy problems where using pre-activations with unshared
biases provably improves the representational power of the
corresponding architectures using post-activations.  In
Figure~\ref{fig:separable}a, a binary classification problem that
cannot be solved by a standard fully connected layer $\R^2 \rightarrow
\R$ can be solved by its DAC counterpart.  In
Figure~\ref{fig:separable}b, a fully connected layer $\R^2 \rightarrow
\R^2$ with unshared biases is able to separate a dataset that is not
separable by a standard fully connected layer $\R^2 \rightarrow \R^2$.

\subsection{Optimal use of parameters for piecewise-linear functions}
\label{sec:params_efficiency}

To support the statement that DAC units allow better use of parameters, we prove that piecewise-linear functions $\R\rightarrow\R$ with $k$ components can be represented with $2k$ parameters by fully connected DAC-enhanced layers, and not less than $3k+1$ parameters by fully connected layers with standard units. In this example, DAC units obtain the best theoretical parameter efficiency.

\begin{mytheorem}
\label{thm:piecewise}
Let $k$ be a positive integer, and let $\text{PL}_k$ be the set of continuous piecewise linear functions $\R\to\R$ consisting of exactly $k$ linear components. Then:
\begin{enumerate}
\item There exists a one-layer fully connected DAC-enhanced neural network $f_k$ representing $\text{PL}_k$. The input is replicated $k$ times, and there are $2k$ parameters.
\item A standard neural network with one layer cannot represent $\text{PL}_k$, with or without input replication.
\item For a standard neural network, the minimal structure needed to represent $\text{PL}_k$ is two layers and $3k+1$ parameters. 
\end{enumerate}
\end{mytheorem}

\begin{proof}\
\begin{enumerate}
\item Let $f_{k,\mathbf{w},\mathbf{b}}$ be a one-layer fully connected neural network with one DAC unit, with weight and bias vector $\mathbf{w}\in\R^k$ and $\mathbf{b}\in\R^k$, respectively, and input $x\in\R$ replicated $k$ times. By Formula~\eqref{eq:dac_equation}, DAC output is
\[
f_{k,\mathbf{w},\mathbf{b}}(x)=\sum_{j=1}^kw_j\varphi(b_j+x),
\]
and $\text{PL}_k=\{f_{k,\mathbf{w},\mathbf{b}} | \mathbf{w},\mathbf{b}\in\R^k\}$.
\item Without input replication, a one-layer standard neural network $\R\to\R$ is just $f(x)=\varphi(wx+b)$. With input replication, the representational power does not change, because
\[
f_{k,\mathbf{w},b}(x)=\varphi(b+\sum_{j=1}^kw_j x)=\varphi(b+(\sum_{j=1}^kw_j)x).
\]
\item By point (2), we need at least two layers with a standard neural network. Since we need to represent $k$ different slopes, we need at least $k$ units in the first layer, for a total of $2k$ parameters ($k$ weights and $k$ biases). The second layer is then a single unit, adding $k$ additional weights and one bias. Thus, a standard neural network needs at least $2k+k+1=3k+1$ parameters to represent the whole $\text{PL}_k$.  
\end{enumerate}
\end{proof}

\begin{remark}
Note that $\text{PL}_k$ depends on exactly $2k$ parameters ($k$ slopes and $k$ intercepts), so that DAC units realize $\text{PL}_k$ with the theoretical minimum number of parameters.
\end{remark}


\clearpage

\section{Additional plots}

\begin{figure}[tb]
  \centering
  \includegraphics[width=0.5\columnwidth, trim=0 0 0 0, clip]{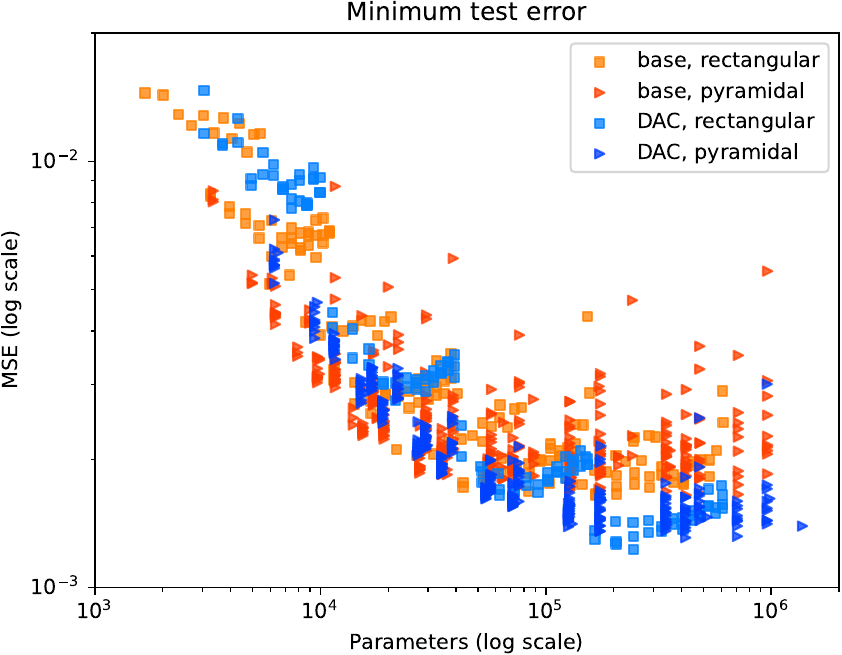}
  \caption{Comprehensive results of 1008 experiments for the SGEMM
    regression task.  Some outliers were removed (see text).
    Fully connected networks with unshared biases (DAC, blue) perform
    worse than networks with shared biases (baseline, orange), for
    number of parameters up to $10^4$, but become generally better for
    number of parameters above $10^5$, also exceeding the best results
    of baseline networks.}
  \label{fig:fc_all_experiments}
\end{figure}

\begin{figure}[ht]
        \centering
        \includegraphics[width=0.5\columnwidth]{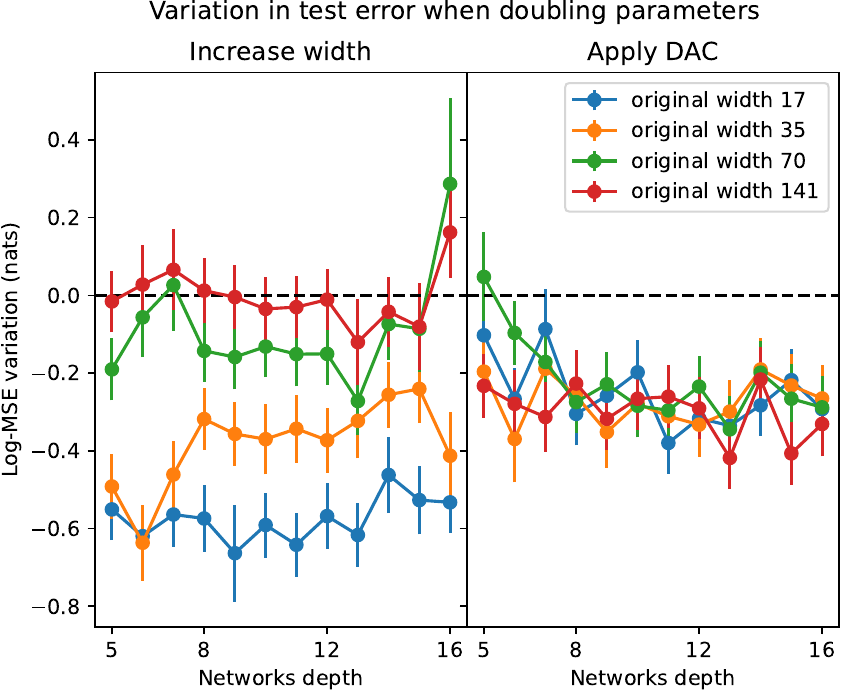}
        \caption{Efficiency analysis of unshared biases for the SGEMM
          regression task.  Rectangular baseline networks were
          compared with models with double the parameters: either by
          adding weights (left) or by making biases unshared (right).
          The resulting variations of the MSE are shown (negative
          means improvement).  Error bars are 95\% confidence
          intervals for the true values.  These plots show no
          dependence on the depth.  These are the original values
          before merging depths that yielded the plot in
          Figure~\ref{fig:fc_double_pars}.}
        \label{fig:doubling_depth_analysis}
\end{figure}

\begin{figure}[ht]
        \centering
        \includegraphics[width=0.5\columnwidth, trim=0 0 0 0, clip]{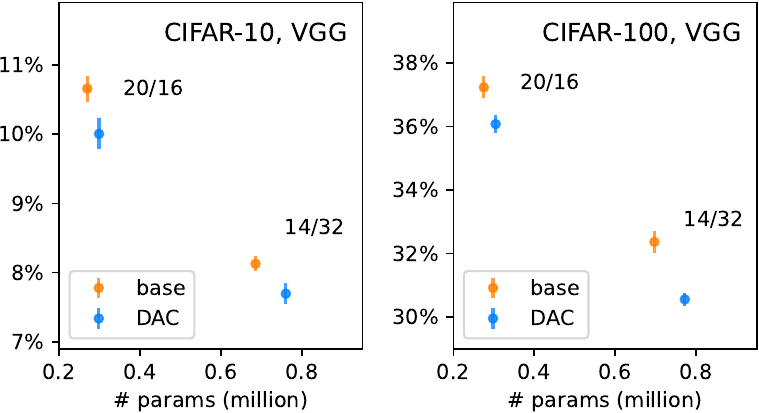}
        \caption{Same as Figure~\ref{fig:vgg} but with model size on
          the horizontal axis.}
        \label{fig:vgg_param}
\end{figure}

\begin{figure}[ht]
        \centering
        \includegraphics[width=0.5\columnwidth, trim=0 0 0 0, clip]{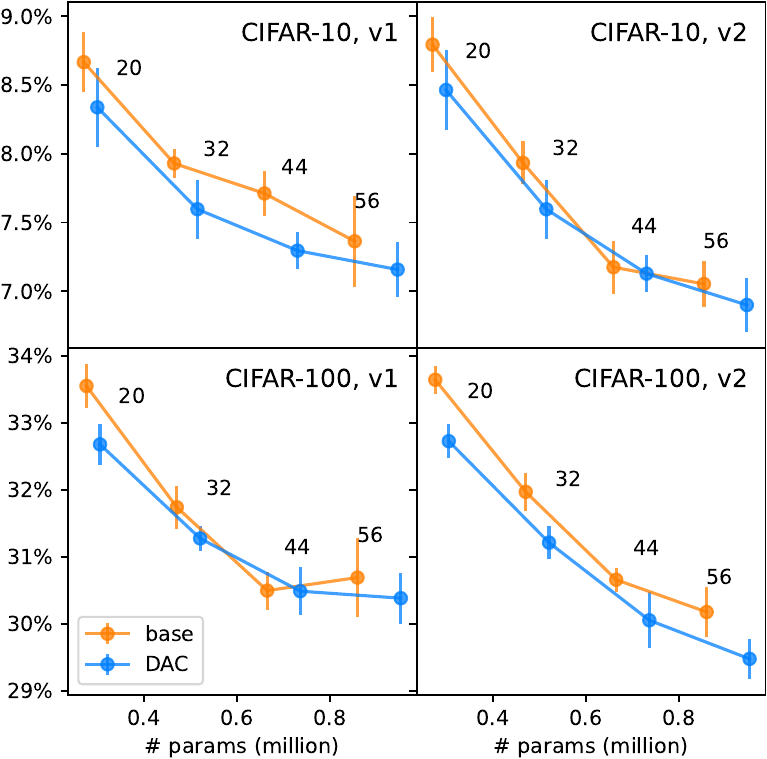}
        \caption{Same as Figure~\ref{fig:comparison_results} but with
          model size on the horizontal axis.}
        \label{fig:comparison_results_param}
\end{figure}


\end{document}